\documentclass[a4paper]{cas-sc}
\usepackage[authoryear,longnamesfirst]{natbib}
\usepackage{caption}
\usepackage{times}
\usepackage{latexsym}
\usepackage[T1]{fontenc}
\usepackage{inconsolata}
\usepackage{graphicx} 
\usepackage[section]{placeins}
\usepackage[ruled,vlined,linesnumbered]{algorithm2e} 
\usepackage{amsmath,amssymb} 
\usepackage{tabularx}
\usepackage{float}
\usepackage[utf8]{inputenc}
\usepackage{microtype}
\usepackage{tcolorbox}
\usepackage{booktabs,multirow}
\usepackage{array}
\usepackage{etoolbox} 
\usepackage{threeparttable}
\usepackage{makecell}
\usepackage{url}
\usepackage[normalem]{ulem}


\newcolumntype{P}[1]{>{\centering\arraybackslash}p{#1}}
\newcolumntype{N}[1]{>{\small}P{#1}}
\usepackage{array}

\AtBeginEnvironment{tabular}{\small}

\newcommand{\upgain}[1]{\textcolor{ForestGreen}{\,$\uparrow$\kern-0.08em#1}}
\newcommand{\downgain}[1]{\textcolor{ForestGreen}{\,$\downarrow$\kern-0.08em#1}}
\newcommand{\upgainsmall}[1]{\textcolor{ForestGreen}{\small\,$\uparrow$\kern-0.08em#1}}
\newcommand{\secondbest}[1]{\uline{#1}}

\tcbuselibrary{skins}

\shorttitle{APCD for Reliable LLM Generation}
\shortauthors{Zheng et al.}

\begin{document}
\let\WriteBookmarks\relax

\title[mode=title]{APCD: Adaptive Path-Contrastive Decoding for Reliable Large Language Model Generation}

\author[1]{Tianyu Zheng}
\author[1]{Hong Wu}[orcid=0000-0002-7019-8046]
\cormark[1]
\ead{hwu@uestc.edu.cn}
\author[1]{Jiaji Zhong}

\affiliation[1]{organization={University of Electronic Science and Technology of China},
    city={Chengdu},
    postcode={611731},
    country={China}}

\cortext[cor1]{Corresponding author}

\begin{abstract}
Reliable text generation is critical for deploying large language models (LLMs) in real-world applications, particularly in high-stakes domains such as medicine. To improve factual reliability, various inference-time methods have been proposed, including logit-level methods that modify token probability distributions and representation-level methods that manipulate intermediate model representations. However, most existing approaches operate on a single decoding trajectory, limiting their ability to explore alternative reasoning paths and making them susceptible to error accumulation. To address this limitation, we propose \textit{Adaptive Path-Contrastive Decoding (APCD)}, an adaptive multi-path contrastive decoding framework that improves factual reliability without model retraining or fine-tuning. APCD comprises two key components: \textit{Entropy-Driven Path Expansion}, which adaptively expands the decoding process only at high-uncertainty decision points, and \textit{Divergence-Aware Path Contrast}, which dynamically regulates contrastive interactions among parallel decoding paths based on their distributional divergence to balance diversity and coherence. We evaluate APCD on four LLM backbones across eight benchmarks spanning both general-domain and medical question answering tasks. Experimental results demonstrate that APCD consistently outperforms strong inference-time baselines in factual accuracy while maintaining competitive inference efficiency. These results demonstrate the robustness and generalizability of APCD across diverse models and tasks, highlighting its effectiveness as a practical multi-path decoding framework for reliable LLM deployment, particularly in high-stakes domains such as medicine. Code is available at https://github.com/zty-king/APCD.

\end{abstract}



\maketitle

\section{Introduction}
Large language models (LLMs) have demonstrated remarkable capabilities across a wide range of natural language tasks owing to their large-scale parameters and strong generalization ability \citep{Language-model}. However, their reliability remains a major concern, particularly in high-stakes domains such as finance \citep{financial-llm}, law \citep{law-llm}, and medicine \citep{medicine-llm1,medicine-llm2}, where hallucinated or factually incorrect outputs may lead to serious consequences. Although Chain-of-Thought (CoT) prompting \citep{CoT1,CoT2,CoT3} improves reasoning transparency, conventional decoding strategies, such as greedy decoding \citep{greedy} and sampling \citep{sample_top_k}, remain vulnerable to early token prediction errors that can propagate and accumulate throughout the generation process.

To improve the factual reliability of LLM generation, numerous inference-time methods have been proposed. These approaches can be broadly categorized into logit-level methods (e.g., CS \citep{CS}, CD \citep{CD}, and CAD \citep{CAD}) and representation-level methods (e.g., ITI \citep{ITI}, DoLa \citep{Dola}, and AD \citep{AD}). While effective, these methods typically operate on a single decoding trajectory and rely on manually designed heuristics, limiting their ability to explore alternative reasoning paths and making them susceptible to error accumulation in multi-step reasoning.

Multi-path decoding methods offer a complementary direction by exploring multiple candidate trajectories during generation. Representative approaches include Beam Search \citep{BeamSearch}, Self-Consistency \citep{self-consistency}, and Best-of-\textit{N} \citep{Best-of-N}. Although these methods improve exploration, they often suffer from premature branching, mode collapse, and insufficient coordination among candidate paths. Diversity Beam Search (DBS) \citep{DBS} explicitly promotes path diversity, but its fixed diversity penalties can compromise semantic coherence. Consequently, developing an adaptive multi-path decoding framework with principled branching and effective inter-path interaction remains an open challenge.

\vspace{4pt}
\noindent\begin{minipage}{\linewidth}
    \centering
    \captionsetup{font=footnotesize, skip=4pt}
    \begin{minipage}[t]{0.48\linewidth}
        \centering
        \includegraphics[width=\linewidth]{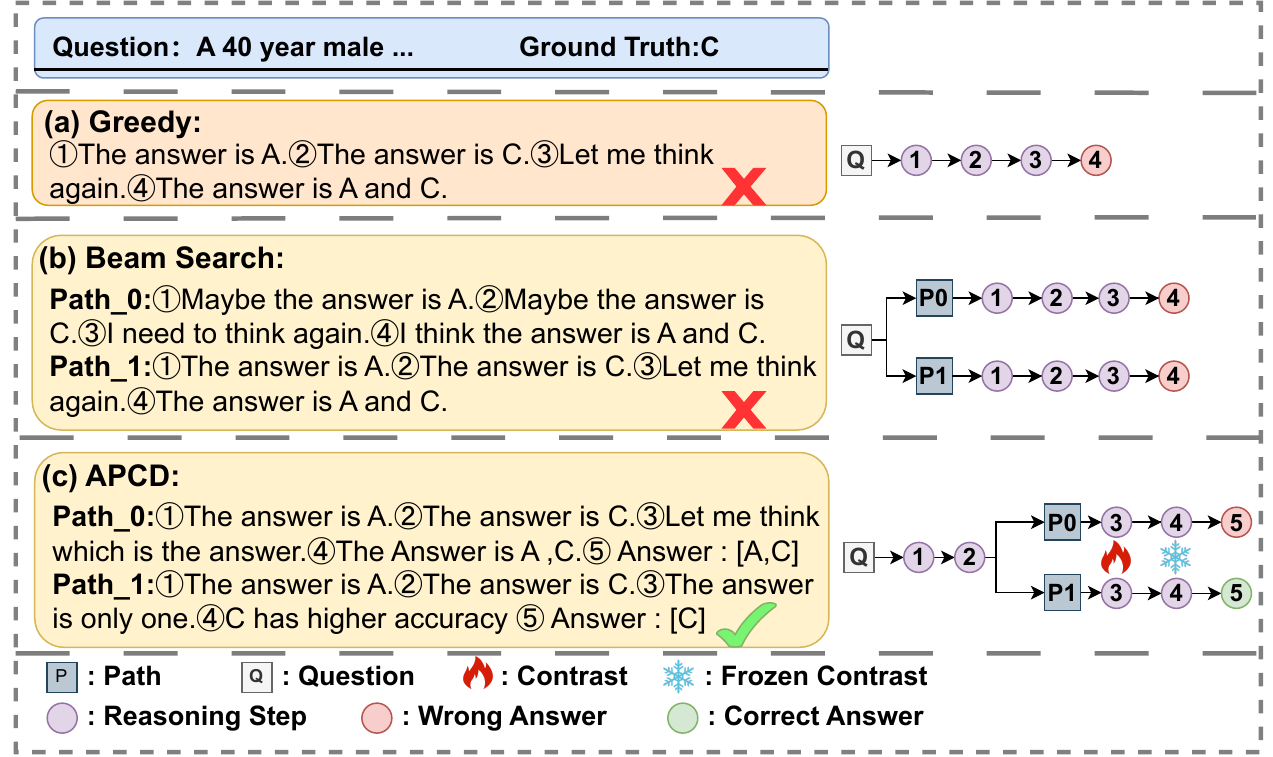}
        \captionof{figure}{Comparison of APCD and traditional approaches on a demonstrative example.: (a) greedy decoding converges to local optima due to early errors; (b) beam search collapses to similar incorrect trajectories; (c) APCD mitigates these issues through adaptive branching and contrastive path interaction, leading to improved reasoning reliability.}
        \label{fig:1}
    \end{minipage}\hfill
    \begin{minipage}[t]{0.48\linewidth}
        \centering
        \includegraphics[width=\linewidth]{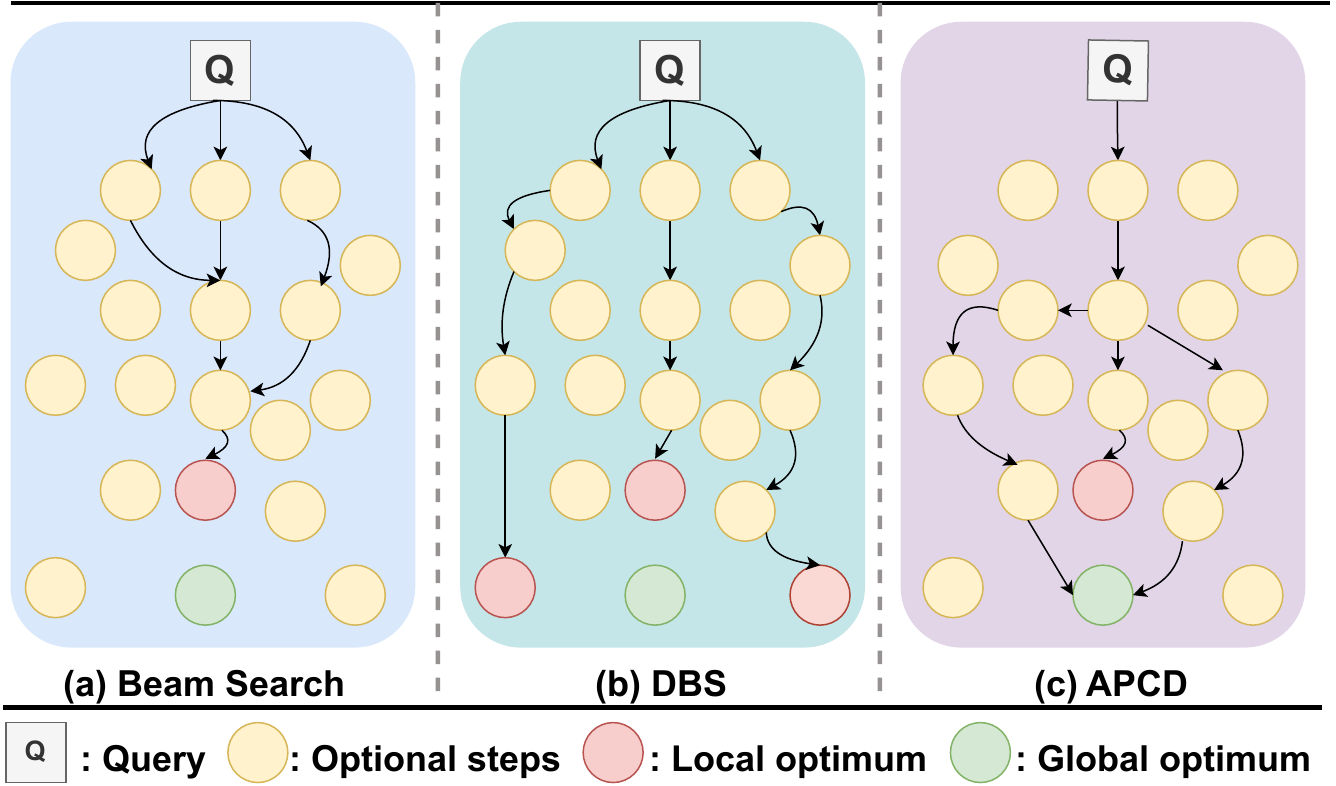}
        \captionof{figure}{Comparison of multi-path decoding behaviors: (a) beam search suffers from trajectory collapse into similar local optima; (b) diversity beam search increases diversity but may degrade reasoning quality due to rigid constraints; (c) APCD achieves both diversity and reliability through adaptive branching and contrastive path interaction.}
        \label{fig:2}
    \end{minipage}
\end{minipage}
\vspace{4pt}

To overcome the limitations of both existing inference-time methods and multi-path decoding approaches, we propose \textit{Adaptive Path-Contrastive Decoding} (APCD), an adaptive multi-path contrastive decoding framework that improves factual reliability without model retraining or fine-tuning (Figure~\ref{fig:1}). APCD consists of two key components. First, \textit{Entropy-Driven Path Expansion} adaptively initiates branching only when predictive uncertainty indicates multiple plausible continuations, enabling efficient exploration while avoiding unnecessary path expansion. Second, \textit{Divergence-Aware Path Contrast} dynamically regulates interactions among parallel decoding paths based on their distributional divergence, promoting diverse yet semantically coherent reasoning trajectories. Finally, APCD selects the output sequence with the lowest perplexity as the final prediction, reflecting overall sequence confidence. As illustrated in Figure~\ref{fig:2}, APCD effectively integrates adaptive multi-path exploration with contrastive path interaction, mitigating mode collapse while preserving semantic coherence and ultimately improving factual reliability.

We evaluate APCD on four LLM backbones (LLaMA-2-7B, LLaMA-2-13B, LLaMA-3-8B, and Qwen-II-Medical-8B) across eight benchmarks, including four general-domain and four medical question answering datasets. Experimental results demonstrate that APCD consistently outperforms strong inference-time baselines and traditional multi-path decoding methods, including CS, CAD, ITI, DoLa, AD, Beam Search and DBS, in factual accuracy while maintaining competitive inference efficiency.

Our main contributions are summarized as follows: 
\begin{itemize} 
\item We propose APCD, an adaptive multi-path contrastive decoding framework that improves the factual reliability of LLMs without model retraining or fine-tuning. 
\item We introduce an entropy-driven path expansion mechanism that adaptively performs multi-path exploration only at high-uncertainty decision points. 
\item We develop a divergence-aware path contrast mechanism that dynamically coordinates parallel decoding paths to balance diversity and semantic coherence. 
\item We conduct extensive experiments on four LLM backbones and eight benchmarks, demonstrating consistent improvements in factual accuracy while maintaining competitive inference efficiency. 
\end{itemize}

\section{Related Works}
\subsection{Inference-time methods}
To improve factuality and mitigate hallucinations in large language models (LLMs), numerous inference-time methods have been proposed without updating model parameters. These approaches can be broadly categorized into \emph{logit-level} and \emph{representation-level} methods.

Logit-level methods improve generation by modifying the output logits or token probability distribution before token selection. Among them, a prominent line of work follows the \emph{contrastive principle}, which enhances generation by contrasting different probability distributions or logits during decoding. Representative approaches include Contrastive Search (CS) \citep{CS}, which mitigates text degeneration by penalizing repetitive tokens, Contrastive Decoding (CD) \citep{CD}, which improves factuality by contrasting the output distributions of expert and amateur models, and Context-aware Decoding (CAD) \citep{CAD}, which enhances faithfulness to the input context by contrasting context-aware and context-agnostic logits.

In contrast, representation-level methods steer generation by manipulating or exploiting intermediate hidden representations. Inference-Time Intervention (ITI) \citep{ITI} amplifies attention heads associated with truthfulness to guide generation, while Decoding by Contrasting Layers (DoLa) \citep{Dola} exploits discrepancies between mature and premature layer predictions to better elicit factual knowledge. Activation Decoding (AD) \citep{AD} further leverages activation patterns from informative layers to improve factual consistency.

More recently, uncertainty-aware decoding methods have incorporated predictive uncertainty into the decoding process. Entropy eNhanced Decoding (END) \citep{END} utilizes entropy estimated from early-layer predictions to refine the final decoding distribution, whereas Uncertainty-Aware Contrastive Decoding (UCD) \citep{UCD} dynamically adjusts the relative contributions of expert and amateur models according to predictive uncertainty. These approaches are closely related to our work in that they exploit uncertainty signals to improve decoding reliability.

\textbf{Our Innovation:} Unlike existing inference-time methods, which typically operate on a single decoding trajectory and often rely on external expert models or predefined intervention strategies, APCD extends the \emph{contrastive principle} underlying existing logit-level decoding methods to an adaptive multi-path decoding framework. Specifically, APCD performs adaptive path expansion based on predictive uncertainty and introduces divergence-aware cross-path contrast, enabling multiple candidate trajectories to interact dynamically according to their distributional divergence. By integrating contrastive reasoning across multiple decoding paths, APCD combines the strengths of contrastive decoding methods and multi-path exploration, leading to more robust and reliable factual generation.

\subsection{Multi-Path Decoding}
Multi-path decoding methods improve generation robustness by exploring multiple candidate trajectories during inference instead of relying on a single decoding path. This paradigm is widely studied to mitigate the limitations of greedy decoding in large language models.

Early approaches such as Beam Search \citep{BeamSearch} maintain a set of high-probability candidates to approximate globally likely outputs. However, Beam Search is prone to mode collapse, where multiple beams converge to similar or identical solutions. Diverse Beam Search (DBS) \citep{DBS}, an earlier extension of beam-based decoding, introduces diversity penalties to encourage exploration; nevertheless, its fixed penalty design is not adaptive across reasoning stages and can degrade semantic coherence in complex generation tasks.

To explore coarser-grained reasoning alternatives, sentence-level and stage-level Beam Search variants \citep{Sentence-leve,Stage_level} extend branching beyond token level. However, these approaches remain limited by early branching decisions and accumulated local optimality.

Sampling-based methods such as Self-Consistency \citep{self-consistency} and Best-of-N \citep{Best-of-N} generate multiple independent trajectories and select outputs via majority voting or scoring mechanisms. While effective, they introduce significant computational overhead and lack mechanisms for correcting reasoning errors during generation \citep{High-reasoning1,High-reasoning2}.

\textbf{Our Innovation:} In contrast to prior multi-path decoding methods that rely on fixed branching rules or heuristic diversity constraints, APCD introduces a reliability-aware adaptive exploration strategy. Path expansion is dynamically triggered based on predictive uncertainty, ensuring that branching occurs only at informative decision points. Moreover, APCD incorporates controlled inter-path interaction to promote effective exploration while preserving reasoning coherence. This leads to a more effective multi-path inference process to improve factual reliability.

\section{Adaptive Path-Contrastive Decoding}
\FloatBarrier

The proposed Adaptive Path-Contrastive Decoding (APCD) framework aims to improve the reliability and interpretability of LLM generation through adaptive multi-path exploration and adaptive path-level contrastive interaction. As illustrated in Figure~\ref{fig:3}, APCD consists of two core components: Entropy-Driven Path Expansion (EDPE) and Divergence-Aware Path Contrast (DAPC). Decoding is initialized with a single path, and uncertainty over the top candidate tokens is continuously monitored by entropy to identify appropriate points for path expansion, enabling exploration only when it is warranted. Once decoding transitions from a single-path to a multi-path regime, Divergence-Aware Path Contrast dynamically adjusts each path's token prediction distribution by contrasting it with those of other active paths. The strength of inter-path interaction is adaptively modulated based on the Jensen--Shannon divergence (JSD) between prediction distributions, promoting effective exploration while avoiding unnecessary interference. Finally, the output is selected as the path with the lowest perplexity.

\vspace{4pt}
\noindent\begin{minipage}{\linewidth}
    \centering
    \includegraphics[width=0.88\linewidth]{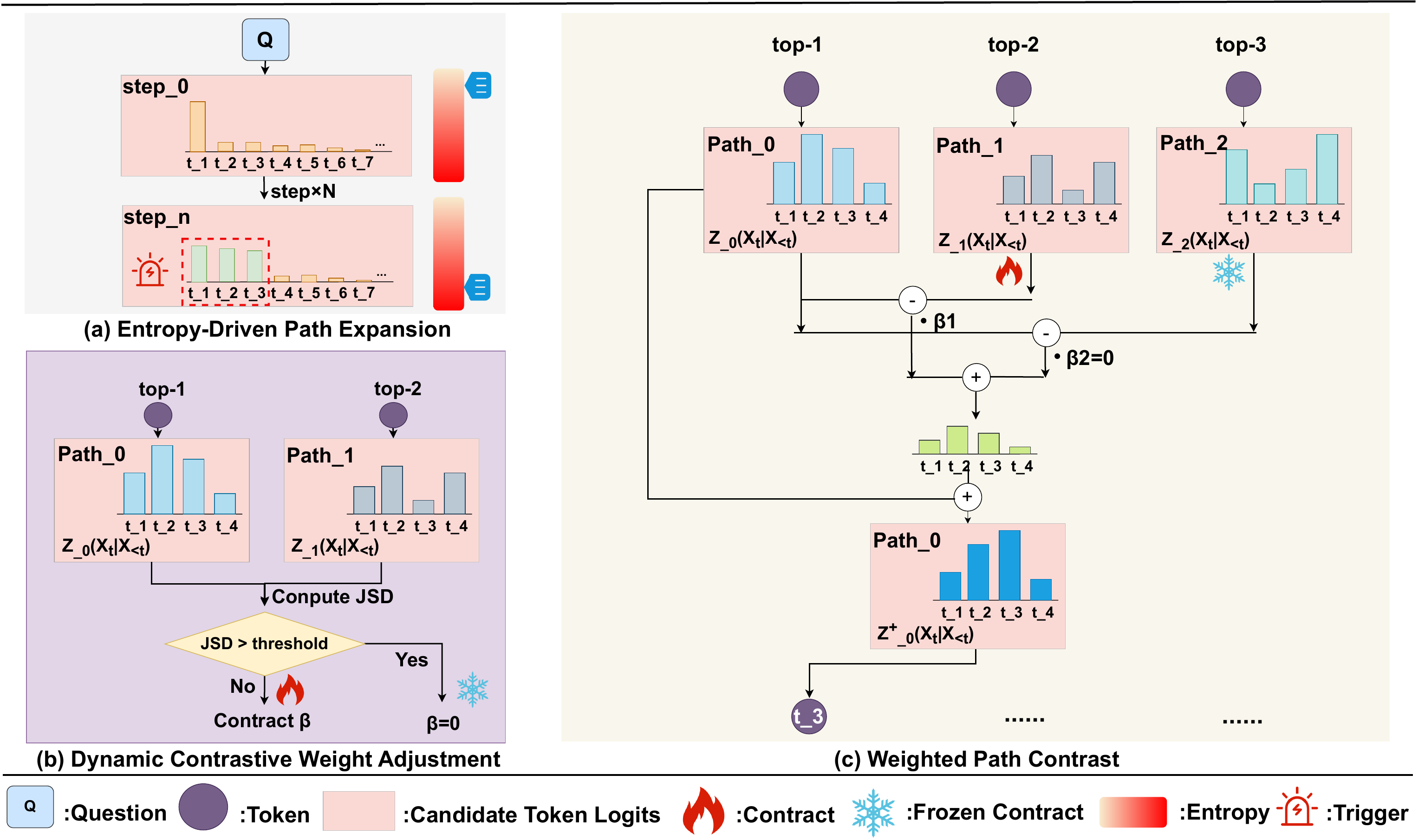}
    \captionof{figure}{APCD Algorithm Flowchart: (a) Entropy-Driven Path Expansion: Multi-path decoding is triggered when entropy exceeds a threshold (indicating comparable top-$k$ likelihoods).(b)-(c) Divergence-Aware Path Contrast:First(b),for each path pair, the contrast weight $\beta$ is adjusted by the JSD between the contrastive paths ($\beta$=0 if it exceeds the threshold).Second(c), The logits of each path are updated by weighted contrasting to the others.}
    \label{fig:3}
\end{minipage}
\vspace{4pt}

\subsection{Entropy-Driven Path Expansion}
We formulate the solution to a query $q$ as a reasoning trajectory $\mathbf{y} = (y_1, \dots, y_L)$. The joint probability of the sequence is modeled autoregressively using the chain rule:
\begin{equation}
P(\mathbf{y} \mid q) = \prod_{t=1}^{L} P(y_t \mid \mathbf{y}_{<t}, q)
\end{equation}

At step $t$, the model projects the context $\mathbf{y}_{<t}$ into logits $\mathbf{z}_t \in \mathbb{R}^M$, where $M$ denotes the size of vocabulary set $\mathcal{V}$, yielding the conditional probability for the token $k$:
\begin{equation}
p_k=P(y_t = k \mid \mathbf{y}_{<t}, q)
= \frac{e^{z_{t,k}}}{\sum_{j=1}^{M} e^{z_{t,j}}}
\end{equation}

Typically, LLMs exhibit a relatively stable phase at the early stage of reasoning, during which the prediction distribution is more concentrated and the entropy is relatively low, indicating high confidence in the selected token. When a single next token dominates the prediction distribution, expanding paths from low-probability tokens often leads to ineffective exploration. Accordingly, path expansion is deferred until the top-$k$ candidates exhibit comparable likelihoods, which can be detected using Shannon entropy. In Chain-of-Thought reasoning, such entropy peaks often correspond to decision points where multiple plausible reasoning steps compete, making exploration particularly valuable \citep{COT-entropy}. Motivated by this observation, we introduce an Entropy-Driven Path Expansion (EDPE) mechanism. Decoding is initialized with a single path and continuously monitors uncertainty over the top-$k$ candidate tokens. When the entropy exceeds a predefined threshold---indicating that multiple top candidates have comparable likelihoods---the decoder selectively expands the single path into multiple paths initialized by these tokens.

To quantify uncertainty, we compute the Shannon entropy over the top-$k$ candidate tokens as:
\begin{equation}
p'_i = \frac{p_i}{\sum_{j=1}^{k} p_j}\quad
H = - \sum_{i=1}^{k} p'_i \ln p'_i
\end{equation}
where $p'_i$ denotes the probability of token $i$ re-normalized over the top-$k$ tokens. The entropy is further normalized as $H' = H / H_{\max}$, where $H_{\max} = \ln k$ is the maximum entropy of a $k$-way categorical distribution. When $H'$ exceeds a threshold $H_\theta$, indicating high uncertainty among the top-$k$ candidates, the decoding process transitions from single-path to $k$-path decoding to enable multi-path exploration. The algorithmic description of EDPE is given in Algorithm~\ref{alg:apcd}.

\begin{algorithm}[t]
\caption{EDPE}
\label{alg:entropy_final}
\KwIn{LLM $\theta$, Multi-path width $k$, Entropy threshold $H_{\theta}$}
\KwOut{Expansion point logits $z_t$, KV cache $\mathrm{KV}_t$, single path $\mathrm{text}$}

\BlankLine
\textbf{Init:} decoding step $t \gets 1$, convergence flag $C \gets \text{True}$, $\mathrm{KV}_0 \gets \text{empty}$, $tk_0 \gets \text{start token}$, $\mathrm{text} \gets \text{empty string}$ \;

\While{$C$}{
    $(tk_{t}, z_{t}, \mathrm{KV}_{t}) \gets \theta(tk_{t-1}, \mathrm{KV}_{t-1})$ \;
    $P_t \gets \operatorname{Softmax}(z_t)$ \;
    $U_t \gets \operatorname{TopK}(P_t, k)$ \;
    $P'_t \gets \operatorname{Normalize}(P_t[U_t])$ \;
    $H_k \gets -\sum_{j \in U_t} P'_t[j] \ln P'_t[j]$ \;
    $H'_k \gets H_k / \ln(k)$ \;
    $\mathrm{text} \gets \mathrm{text}+tk_{t}$ \;
    \If{$H'_k \ge H_{\theta}$}{
        $C \gets \text{False}$\;
        \textbf{break}\;
    }
    $t \gets t + 1$ \;
}
\Return{$z_t, \mathrm{KV}_t$}, $\mathrm{text}$\;
\end{algorithm}

\subsection{Divergence-Aware Path Contrast}
Once decoding enters the multi-path regime, APCD applies a Divergence-Aware Path Contrast (DAPC) mechanism to promote effective exploration while preserving reasoning coherence. This mechanism adjusts each path's token prediction distribution by contrasting it with those of other active paths to highlight the distinctive information across the paths. Inspired by contrastive decoding\citep{CD}, the logits of each path are updated as:

\begin{equation}
\label{eq:z_logits}
\tilde{z}_i^b = z_i^b + \sum_{b' \neq b} \beta_{b,b'} \cdot (z_i^b - z_i^{b'}), \quad i \in \mathcal{V}^{b}_{\text{valid}}
\end{equation}
where $z_i^b$ and $z_i^{b'}$ denotes the logits of candidate token $i$ for path $b$ and $b'$. Contrastive weight $\beta_{b,b'}$ controls the strength of contrastive interaction between the two paths. The update is applied only to the vocabulary subset $\mathcal{V}^{b}_{\text{valid}}$ to improve efficiency:
\begin{equation}
\label{eq:Valid}
\mathcal{V}^{b}_{\text{valid}} = \left\{ i \mid z_i^b \geq \max_{j \in \mathcal{V}} (z_j^b) + \log \alpha \right\}
\end{equation}
where $\alpha \in (0,1)$ controls the size of the subset. This restriction confines updates to tokens near the peak logits, substantially reducing computational overhead while preserving informative alternatives. This logit updating formulation preserves the information of each path while incorporating relevant signals from other paths, enabling effective multi-path exploration.

When the divergence between the prediction distributions of a pair of paths becomes large, further enlarging it may harm reasoning coherence. Therefore, the contrastive weight is dynamically adjusted based on the divergence between paths. Specifically, the contrastive weight between paths $(b, b')$ is defined as
\begin{equation}
\label{eq:beta}
\beta_{b,b'} = \beta_{\text{max}} \cdot \max \left( 0,1-\frac{JSD(\hat{\mathbf{p}}^b \| \hat{\mathbf{p}}^{b'})}{\delta \log 2} \right)
\end{equation}
where $JSD(\cdot \| \cdot)$ denotes the Jensen--Shannon divergence, normalized to $[0,1]$ by $\log 2$, $\beta_{max}$ denotes the initial value of the pairwise path-contrastive weight. Here, $\hat{\mathbf{p}}^b$ and $\hat{\mathbf{p}}^{b'}$ are the token probability distributions of paths $b$ and $b'$, computed over the union of $\mathcal{V}^{b}_{\text{valid}}$ and $\mathcal{V}^{b'}_{\text{valid}}$. $\delta$ is the divergence threshold: when the normalized JSD exceeds $\delta$, inter-path interaction is terminated by setting $\beta_{b,b'} = 0$; otherwise, smaller divergence leads to stronger interaction.

Overall, the path-contrast mechanism adaptively regulates multi-path exploration based on inter-path divergence. In early multi-path decoding stages, contrastive interaction encourages paths to explore distinct trajectories and prevents premature path collapse. As paths diverge, inter-path interaction is gradually attenuated and eventually halted, allowing coherent reasoning trajectories to evolve without forced separation.

\subsection{Final Solution Selection}
\label{alg:ppl}
To evaluate the reliability of each generated trajectory, we adopt perplexity as a global measure of sequence coherence and model confidence. Specifically, the perplexity of a trajectory is defined as:
\begin{equation}
\label{eq:ppl}
\text{ppl}(\mathbf{y}) = \exp \left( - \frac{1}{L} \sum_{t=1}^{L} \log P(y_t \mid \mathbf{y}_{<t}, q) \right)
\end{equation}
where $L$ denotes the trajectory length. 

During decoding, trajectories whose perplexity exceeds a predefined threshold $\tau$ are pruned, as they are considered less likely to represent reliable reasoning paths. After generation, the final output is selected as the remaining trajectory with the lowest perplexity.

The overall procedure of APCD is summarized in Algorithm~\ref{alg:apcd}.

\begin{algorithm}[t]
\caption{APCD}
\label{alg:apcd}
\KwIn{
  LLM $\theta$, multi-path width $k$, threshold $\alpha$, $\delta$, $H_{\theta}$, initial contrast weight $\beta_{\max}$, max steps $T$,Vocab Size $M$
}
\KwOut{Generated best $\mathrm{text}^*$}
\tcp{Single-path Decoding}
$(\mathbf{z}^{t_c},{\mathrm{KV}}^{t_c},\mathrm{text}^{t_c}) \gets \text{EDPE } (\theta,H_{\theta},k)$\;
$\{ \mathrm{tk}_i^{t_c} \}_{i=1}^k \leftarrow \operatorname{TopK}(\mathbf{z}^{t_c}, k)$, $t \leftarrow t_c+1$ \;
$\{\mathrm{KV}_i^{t_c}\}_{i=1}^k \leftarrow \operatorname{Replicate}(\mathrm{KV}^{t_c}, k)$\;
$ \{\mathrm{text}_i\}_{i=1}^k \leftarrow \operatorname{Replicate}(\mathrm{text}^{t_c})$ \;
\tcp{Multi-path Decoding}
\While{$t \le T$}{
  \For{$b,b' \in \{1,\dots,k\},\, b\ne b'$}{
    $ (\,\mathbf{z}^t_b,\ \mathrm{KV}^t_b\,) \leftarrow \theta(\,\text{tk}_{b}^{t-1},\ \mathrm{KV}_{b}^{t-1}\,)$\;
    Compute $\mathcal{V}^{b'}_{\text{valid}}$, $\mathcal{V}^{b}_{\text{valid}}$ via Eq.~\ref{eq:Valid}\;
    Compute $\beta_{b,b'}$ via Eq.~\ref{eq:beta}\;
  }

  \For{$b = 1,\dots,k$}{ 
    Compute $\mathbf{\tilde z}^t_b  = [\tilde z_i^b]_{i \in \mathcal{V}^{b}_{\text{valid}}}$ via Eq.~\ref{eq:z_logits} \; 
    $\text{tk}^{t}_b \leftarrow \arg\max(\operatorname{Softmax}(\mathbf{\tilde z}^t_b))$\;
    $\mathrm{text}_b \leftarrow \mathrm{text}_b+\text{tk}^{t}_b$\
  }
}
$\text{ppl}_b \gets \text{Compute ppl via Eq.~\ref{eq:ppl}} $\;
$\mathrm{text}^* \leftarrow \arg\min_{b \in \{1,\dots,k\}}\ \mathrm{ppl}_b$\;
\Return{$\mathrm{text}^*$}\;
\end{algorithm}

\section{Empirical Evaluation}
In this section, we evaluate the effectiveness of APCD on both general-domain question answering tasks and domain-specific medical question answering tasks.

\subsection{Experiment Setup}
\textbf{Datasets:} We evaluate our method on two categories of general-domain benchmarks: truthfulness-oriented and knowledge-seeking datasets. For truthfulness evaluation, we use TruthfulQA \citep{truthfulqa} and conduct open-ended text generation. For knowledge-seeking tasks, we consider commonly-used question answering benchmarks including TriviaQA \citep{triviaqa}, HotpotQA \citep{hotpotqa}, and Natural Questions (NQ) \citep{naturalquestions}.

To further assess performance in domain-specific and long-form generation scenarios, we conduct experiments on four medical QA benchmarks: MedQA \citep{medqa}, MedMCQA \citep{medmcqa}, MedB-4 \citep{MedB-4}, and NEJM.

\textbf{Evaluation Metrics:} For TruthfulQA, we follow the evaluation protocol of \citep{Dola,ITI} and use the OpenAI API model \citep{openai} (GPT-4o-mini) to assess the truthfulness (Truth), informativeness (Info), their combined score ($\text{Truth*Info}$), and the rejection rate (Reject) of generated responses. For TriviaQA, HotpotQA, and NQ, we adopt the standard evaluation protocol of \citep{triviaqa} and report two widely used factual correctness metrics: Exact Match (EM) and F1 score. For MedQA, MedMCQA, MedB-4, and NEJM, we evaluate factual accuracy using Strict Match (SM) accuracy, which requires exact adherence to the expected answer format, and Fuzzy Match (FM) accuracy, which uses GPT-4o-mini to extract and evaluate the final answer from generated responses.

\textbf{Models \& Baselines:} For general-domain tasks, we use LLaMA-2-7B-Chat \citep{llama2}, LLaMA-2-13B-Chat \citep{llama2}, and LLaMA-3-8B-Instruct \citep{llama3}. For medical-domain tasks, we use II-Medical-8B-1706 \citep{Qwen-II-Medical-8B-1706}, an open-source medical LLM built on the Qwen3-8B backbone developed by Intelligent Internet\footnote{\url{https://huggingface.co/Intelligent-Internet/II-Medical-8B-1706}}.

For general-domain tasks, we compare APCD against several strong inference-time methods, including: (1) Context-Aware Decoding (CAD), (2) Inference-Time Intervention (ITI), (3) DoLa, (4) Activation Decoding (AD), (5) Entropy-Enhanced Decoding (END), and (6) Uncertainty-Aware Contrastive Decoding (UCD).

For medical-domain tasks, we only compare with HuggingFace implementation of Dola based on Qwen model with  upper/lower-layer settings (high/low), because it is not easy to transfer other inference-time methods to Qwen model. In addition, we also compare with 7) Sampling-based decoding and two traditional multi-path decoding methods: 8) Beam Search and 9) Diverse Beam Search (DBS).

\textbf{Implementation Details:} For APCD, we set $k=3$ to balance effectiveness and computational efficiency, and use an entropy threshold $H_{\theta}=0.95$ to ensure branching occurs only under high uncertainty. We set the initial contrastive weight $\beta_{\max}=0.5$ and the divergence threshold $\delta=0.5$ for computing the contrastive weight in Eq.~\ref{eq:beta}. To improve inference efficiency, we set $\alpha = 0.2$ in Eq.~\ref{eq:Valid} to filter low-probability tokens, and apply a perplexity threshold $\tau = 2$ to prune trajectories during decoding. Overall, these hyperparameter settings are stable across different tasks, reducing the need for extensive tuning in practical applications.

In general-domain experiments, we limit the maximum number of generated tokens to 100 due to the short-text nature of the tasks. For baseline inference-time methods, we follow their original implementations, except for ITI, for which we directly use the LLaMA models with ITI implementation\footnote{\url{https://huggingface.co/jujipotle}}. For UCD, we use LLaMA-2-7B-Chat as the base model and LLaMA-2-13B-Chat as the expert model.

In medical-domain experiments, we set the maximum number of generated tokens to 4096 to accommodate the long-form nature of the tasks and ensure sufficient CoT reasoning. We set $num\_beams = 3$ for Beam Search, and for DBS we use $num\_beam\_groups = 3$, $num\_beams = 3$, and $diversity\_penalty = 0.5$, ensuring all multi-path methods use the same number of paths/beams. For sampling-based decoding, we use $top\_k=40$ and $top\_p=0.9$, following common practice.

All experiments are conducted on four NVIDIA Tesla V100 32GB GPUs. Dataset- and model-specific prompts are provided in Appendix~\ref{appendix:Prompt_restore}.

\subsection{Main Results}
\begin{table}[H]
  \centering
  \caption{The experimental results on four general-domain benchmarks (with the metric unit as $\times 10^{-2}$): TruthfulQA, TriviaQA, HotPotQA and Natural Questions. Best-performing results per model size and dataset are highlighted in bold, and up-arrows indicate improvement over greedy decoding.}
  \label{tab:1}
  \resizebox{\linewidth}{!}{%
  \begin{tabular}{l|cc cc cc cc|cc cc|cc cc|cc cc}
    \toprule
    Model & \multicolumn{8}{c|}{\textbf{TruthfulQA}} & \multicolumn{4}{c|}{\textbf{TriviaQA}} & \multicolumn{4}{c|}{\textbf{HotPotQA}} & \multicolumn{4}{c}{\textbf{NQ}} \\
    \cmidrule{2-9}\cmidrule{10-13}\cmidrule{14-17}\cmidrule{18-21}
    & \multicolumn{2}{c}{\%Truth$\uparrow$} & \multicolumn{2}{c}{\%Info$\uparrow$} & \multicolumn{2}{c}{\%T$*$I$\uparrow$} & \multicolumn{2}{c|}{\%Reject$\downarrow$} & \multicolumn{2}{c}{EM$\uparrow$} & \multicolumn{2}{c|}{F1$\uparrow$} & \multicolumn{2}{c}{EM$\uparrow$} & \multicolumn{2}{c|}{F1$\uparrow$} & \multicolumn{2}{c}{EM$\uparrow$} & \multicolumn{2}{c}{F1$\uparrow$} \\
    \midrule
    Llama-2-7b-chat
      & 54.3 & & 81.4 & & 39.5 & & 19.5 & & 58.2 & & 54.8 & & 25.5 & & 23.4 & & 34.2 & & 31.2 & \\
    \quad + ITI
      & 56.4 & & 79.7 & & 40.2 & & 20.7 & & 57.6 & & 52.2 & & 25.8 & & 22.6 & & 33.5 & & 28.1 & \\
    \quad + CAD
      & 54.3 & & 80.2 & & 39.7 & & 19.4 & & 54.0 & & 48.0 & & 24.3 & & 18.8 & & 30.0 & & 25.5 & \\
    \quad + Dola
      & 54.5 & & 81.0 & & 39.2 & & 19.8 & & \secondbest{60.2} & & \secondbest{56.8} & & \secondbest{27.4} & & \secondbest{25.4} & & \secondbest{35.2} & & \secondbest{32.2} & \\
    \quad + AD
      & 54.8 & & 85.0 & & 42.6 & & 16.1 & & 58.6 & & 55.5 & & 26.1 & & 24.2 & & 34.8 & & 31.6 & \\
    \quad + Dola+AD
      & 54.7 & & 84.3 & & 42.1 & & 16.3 & & 58.5 & & 55.5 & & 26.3 & & 24.3 & & 34.9 & & 31.7 & \\
    \quad + End
      & 54.8 & & \textbf{88.0} & & \secondbest{43.2} & & \textbf{9.6} & & 58.6 & & 55.6 & & 25.4 & & 23.8 & & 34.6 & & 31.7 & \\
    \quad + \textbf{Ours}
      & \textbf{56.4} & \textbf{\upgain{2.1}} & \secondbest{85.4} & \textbf{\upgain{4.0}} & \textbf{45.3} & \textbf{\upgain{5.8}} & \secondbest{13.3} & \textbf{\downgain{6.2}}
      & \textbf{60.8} & \textbf{\upgain{2.6}} & \textbf{59.5} & \textbf{\upgain{4.7}} & \textbf{28.0} & \textbf{\upgain{2.5}} & \textbf{28.4} & \textbf{\upgain{5.0}} & \textbf{36.0} & \textbf{\upgain{1.8}} & \textbf{33.7} & \textbf{\upgain{2.5}} \\
    \midrule
    Llama-2-13b-chat
      & 55.4 & & 84.6 & & 44.0 & & 15.8 & & 66.2 & & 62.7 & & 37.8 & & 32.6 & & 36.9 & & 33.8 & \\
    \quad + ITI
      & 57.1 & & 84.9 & & 44.8 & & 14.9 & & 60.9 & & 58.8 & & 24.8 & & 24.4 & & 31.3 & & 26.9 & \\
    \quad + CAD
      & 52.1 & & 85.6 & & 42.6 & & 15.6 & & 62.0 & & 57.8 & & 26.6 & & 23.7 & & 30.0 & & 25.0 & \\
    \quad + Dola
      & \textbf{59.2} & & 80.2 & & 42.8 & & 20.5 & & \secondbest{67.4} & & \secondbest{64.1} & & \secondbest{39.6} & & 36.8 & & \secondbest{39.6} & & 36.8 & \\
    \quad + AD
      & 56.9 & & 88.7 & & 48.1 & & 11.4 & & 66.8 & & 64.0 & & 39.2 & & \secondbest{37.0} & & 39.2 & & \secondbest{37.0} & \\
    \quad + Dola+AD
      & \secondbest{58.7} & & 81.2 & & 43.4 & & 19.6 & & 67.0 & & 64.0 & & 38.8 & & 36.8 & & 38.8 & & 36.8 & \\
    \quad + End
      & 57.1 & & \textbf{90.4} & & \secondbest{48.2} & & \textbf{7.2} & & 66.7 & & 63.2 & & 37.5 & & 33.0 & & 37.3 & & 34.2 & \\
    UCD(+ 7B-Chat)
      & 55.9 & & 86.8 & & 48.0 & & 12.5 & & 62.2 & & 57.7 & & 35.5 & & 31.7 & & 32.6 & & 30.1 & \\
    \quad + \textbf{Ours}
      & 57.5 & \textbf{\upgain{2.1}} & \secondbest{88.8} & \textbf{\upgain{4.2}} & \textbf{48.8} & \textbf{\upgain{4.8}} & \secondbest{11.2} & \textbf{\downgain{4.6}}
      & \textbf{68.0} & \textbf{\upgain{1.8}} & \textbf{66.7} & \textbf{\upgain{4.0}} & \textbf{40.5} & \textbf{\upgain{2.7}} & \textbf{39.0} & \textbf{\upgain{6.4}} & \textbf{41.0} & \textbf{\upgain{4.1}} & \textbf{39.5} & \textbf{\upgain{5.7}} \\
    \midrule
    Llama-3-8b-it
      & 60.3 & & 75.0 & & 39.8 & & 19.9 & & 65.4 & & 62.1 & & 33.1 & & 28.3 & & 37.9 & & 33.5 & \\
    \quad + ITI
      & \textbf{66.4} & & 83.2 & & \textbf{52.6} & & \textbf{16.5} & & 53.1 & & 50.9 & & 21.7 & & 20.6 & & 26.5 & & 21.6 & \\
    \quad + CAD
      & 58.4 & & 76.3 & & 39.2 & & 18.9 & & 63.5 & & 59.4 & & 30.1 & & 26.5 & & 36.5 & & 33.0 & \\
    \quad + Dola
      & 60.6 & & 75.0 & & 40.5 & & 19.9 & & \secondbest{66.8} & & \secondbest{63.1} & & \secondbest{33.6} & & \secondbest{28.8} & & \secondbest{39.0} & & \secondbest{34.5} & \\
    \quad + AD
      & 60.2 & & 74.4 & & 39.9 & & 22.9 & & 66.4 & & 62.5 & & 33.4 & & \secondbest{28.8} & & 38.4 & & 34.0 & \\
    \quad + Dola+AD
      & 60.3 & & 74.3 & & 40.1 & & 22.7 & & 66.5 & & 62.5 & & 33.3 & & 28.6 & & 38.5 & & 34.1 & \\
    \quad + End
      & 60.9 & & \secondbest{83.4} & & 46.2 & & 17.4 & & 64.1 & & 60.9 & & 32.4 & & 28.3 & & 38.5 & & 33.6 & \\
    \quad + \textbf{Ours}
      & \secondbest{63.0} & \textbf{\upgain{2.7}} & \textbf{83.7} & \textbf{\upgain{8.7}} & \secondbest{49.3} & \textbf{\upgain{9.5}} & \secondbest{17.2} & \textbf{\downgain{2.7}}
      & \textbf{68.1} & \textbf{\upgain{2.7}} & \textbf{66.7} & \textbf{\upgain{4.6}} & \textbf{35.2} & \textbf{\upgain{2.1}} & \textbf{33.0} & \textbf{\upgain{4.7}} & \textbf{40.7} & \textbf{\upgain{2.8}} & \textbf{38.0} & \textbf{\upgain{4.5}} \\
    \bottomrule
  \end{tabular}%
  }
\end{table}

\begin{table}[htbp]
  \centering
  \caption{The experimental results on four medical benchmarks (with the metric unit as $\times 10^{-2}$): NEJM, MedMCQA, MedQA and MedB-4. Best-performing results per dataset are highlighted in bold, and up-arrows indicate improvement over greedy decoding.}
  \label{tab:2}
  \resizebox{\linewidth}{!}{%
  \begin{tabular}{l|cc cc|cc cc|cc cc|cc cc}
    \toprule
    Model & \multicolumn{4}{c|}{\textbf{NEJM}} & \multicolumn{4}{c|}{\textbf{MedMCQA}} & \multicolumn{4}{c|}{\textbf{MedQA}} & \multicolumn{4}{c}{\textbf{MedB-4}} \\
    \cmidrule{2-5}\cmidrule{6-9}\cmidrule{10-13}\cmidrule{14-17}
    & \multicolumn{2}{c}{SM} & \multicolumn{2}{c|}{FM} & \multicolumn{2}{c}{SM} & \multicolumn{2}{c|}{FM} & \multicolumn{2}{c}{SM} & \multicolumn{2}{c|}{FM} & \multicolumn{2}{c}{SM} & \multicolumn{2}{c}{FM}\\
    \midrule
    II-Medical-8B-1706
      & 67.7 & & 71.5 & & 67.9 & & 71.0 & & 79.2 & & 82.8 & & 68.4 & & 75.6 & \\
    \quad + Sample (top $k$ + top $p$)
      & 67.2 & & 73.6 & & 72.0 & & 74.0 & & 80.1 & & 85.2 & & 70.8 & & 77.6 & \\
    \quad + CS
      & 72.5 & & 75.3 & & 71.9 & & 73.8 & & 83.4 & & 85.8 & & 75.2 & & 78.1 & \\
    \quad + Beam Search
      & 67.2 & & 73.2 & & 68.9 & & 71.7 & & 79.9 & & 83.1 & & 67.2 & & 72.7 & \\
    \quad + DBS
      & 65.1 & & 72.0 & & 64.2 & & 71.7 & & 72.3 & & 82.0 & & 62.0 & & 75.0 & \\
    \quad + DoLa (high)
      & 11.2 & & 11.5 & & 17.4 & & 17.5 & & 12.3 & & 12.7 & & 8.1 & & 8.1 & \\
    \quad + DoLa (low)
      & 70.3 & & 73.1 & & 69.0 & & 72.9 & & 76.7 & & 84.5 & & 70.1 & & 74.0 & \\
    \quad + \textbf{APCD(Ours)}
      & \textbf{76.6} & \textbf{\upgainsmall{8.9}} & \textbf{78.0} & \textbf{\upgainsmall{6.5}}
      & \textbf{75.3} & \textbf{\upgainsmall{7.4}} & \textbf{76.9} & \textbf{\upgainsmall{5.9}}
      & \textbf{85.0} & \textbf{\upgainsmall{5.8}} & \textbf{86.7} & \textbf{\upgainsmall{3.9}}
      & \textbf{76.0} & \textbf{\upgainsmall{7.6}} & \textbf{78.3} & \textbf{\upgainsmall{2.7}} \\
    \bottomrule
  \end{tabular}%
  }
\end{table}

The experimental results on general-domain tasks are reported in Table~\ref{tab:1}, including open-ended TruthfulQA and knowledge-seeking datasets. For TruthfulQA, our method improves prediction accuracy while preserving informativeness. Compared with greedy decoding and prior methods including ITI, CAD, DoLa, and AD, APCD achieves consistent improvements across all three models. Specifically, compared to greedy decoding, APCD increases $\text{Truth*Info}$ ($\text{T*I}$) by 5.8, 4.8, and 9.5 points for Llama-2-7b-chat, Llama-2-13b-chat, and Llama-3-8b-instruct, respectively. Meanwhile, the $\text{Reject}$ metric decreases by 6.2, 4.6, and 2.7 for the corresponding models. Notably, although DoLa on Llama-2-13b-chat improves accuracy at the cost of lower informativeness, APCD achieves a better balance: it attains a high Truth score of 57.5 while maintaining high informativeness ($\text{Info}$ = 88.8), demonstrating its ability to generate responses that are both factual and sufficiently informative. Despite ITI’s marginal advantage on Llama-3-8b-instruct, our method achieves superior overall performance.

For knowledge-seeking datasets, our method achieves the best performance across all metrics and models. Specifically, relative to greedy decoding on Llama-2-7b-chat, our method improves EM by 2.6 points and F1 by 4.7 on TriviaQA; by 2.5 EM and 5.0 F1 on HotPotQA; and by 1.8 EM and 2.5 F1 on NQ. Similar gains are also observed on LLaMA-2-13B-Chat and LLaMA-3-8B-Instruct. These results indicate that APCD improves factual accuracy while maintaining high answer completeness.

The experimental results on medical-domain tasks are reported in Table~\ref{tab:2}. The results show that APCD outperforms greedy decoding and other comparison methods across all metrics on NEJM, MedMCQA, MedQA, and MedB-4, exhibiting reliable performance in high-risk medical long-form generation. Specifically, compared to greedy decoding, our method achieves consistent improvements across NEJM (SM: +8.9, FM: +6.5), MedMCQA (+7.4, +5.9), MedQA (+5.8, +3.9), and MedB-4 (+7.6, +2.7). In addition to improving accuracy, APCD produces more structured and coherent outputs (according to SM), which facilitates downstream evaluation and reduces post-processing overhead.

\subsection{Ablation Study}
\begin{table}[htbp]
  \centering
  
  \caption{Ablation study of APCD, demonstrating that Entropy-Driven Path Expansion and Divergence-Aware Path Contrast are crucial for preserving model factual accuracy.}
  \label{tab:3}
  \begin{tabular}{l|P{1.4cm} P{1.4cm} P{0.64cm} P{0.64cm} P{0.64cm} P{0.64cm} P{0.64cm} P{0.64cm} P{0.64cm} P{0.64cm}}
    \toprule
    Model & EDPE & DAPC &
      \multicolumn{2}{c}{\textbf{NEJM}} &
      \multicolumn{2}{c}{\textbf{MedMCQA}} &
      \multicolumn{2}{c}{\textbf{MedQA}} &
      \multicolumn{2}{c}{\textbf{MedB-4}}\\
    \cmidrule{4-5}\cmidrule{6-7}\cmidrule{8-9}\cmidrule{10-11}
    & & & SM & FM & SM & FM & SM & FM & SM & FM\\
    \midrule
    \multirow{4}{*}{II-Medical-8B-1706}
      & $\times$ & $\times$ & 67.7 & 71.5 & 67.9 & 71.0 & 79.2 & 82.8 & 68.4 & 75.6\\
      & $\checkmark$ & $\times$ & 76.3 & 77.8 & 73.9 & 75.1 & 84.7 & 86.3 & 75.0 & 77.9\\
      & $\times$ & $\checkmark$ & 65.7 & 71.5 & 61.6 & 70.3 & 71.2 & 82.5 & 74.0 & 76.3\\
      & $\checkmark$ & $\checkmark$ &
        \textbf{76.6} & \textbf{78.0} &
        \textbf{75.3} & \textbf{76.9} &
        \textbf{85.0} & \textbf{86.7} &
        \textbf{76.0} & \textbf{78.3} \\
    \bottomrule
  \end{tabular}%
  
\end{table}

We conducted an ablation study on medical-domain tasks by systematically disabling EDPE or DAPC to evaluate the contribution of each component, and report the results in Table~\ref{tab:3}.

We first disable Entropy-Driven Path Expansion (EDPE) by setting the entropy threshold $H_{\theta}$ to 0, forcing decoding to branch at the initial step. This modification leads to a noticeable performance drop across all benchmarks, particularly on MedMCQA, where the SM score decreases by 13.7 points and the FM score decreases by 6.6 points. This result indicates that EDPE is essential for maintaining informative and stable branches in multi-path decoding.

Next, we keep EDPE and disable Divergence-Aware Path Contrast (DAPC) by setting the initial contrastive weight $\beta_{max}$ to 0, making each reasoning path independent. On MedMCQA, the SM score decreases by 1.4 (from 75.3 to 73.9), and the FM score decreases by 1.8 (from 76.9 to 75.1). Similar trends are observed across other datasets. This experiment highlights the importance of incorporating contrastive interactions to correct and refine ongoing reasoning paths.

These ablation results show that both EDPE and DAPC play critical roles in preserving the model’s factual accuracy and reliability.

\subsection{Comparison of Final Solution Selection Strategies}
Our APCD adopts the lowest-perplexity criterion as the default final solution selection strategy due to its simplicity and efficiency. We compare it with two alternative strategies: majority voting and an external verifier. Majority voting selects the most frequent answer option; in our implementation, it falls back to the lowest-perplexity candidate in cases of ties or extraction failures. The external verifier uses GPT-4o-mini to evaluate and select the best response from three generated candidates.

The results in Table~\ref{tab:10} show that the lowest-perplexity strategy already captures most of the achievable gain: APCD(ppl) improves over greedy decoding by 8.9 SM points on NEJM and 7.6 on MedB-4. Majority voting yields only marginal additional improvement, while the verifier serves as an upper-bound reference.

\begin{table}[htbp]
  \centering
  
  \caption{Analysis of final solution selection strategies for APCD.}
  \label{tab:10}
  \begin{tabular}{l|P{2.7cm}P{2.7cm}P{2.7cm}P{2.7cm}}
    \toprule
    Model &
      \multicolumn{2}{c}{\textbf{NEJM}} &
      \multicolumn{2}{c}{\textbf{MedB-4}} \\
    \cmidrule{2-3}\cmidrule{4-5}
    & SM & FM & SM & FM\\
    \midrule
    II-Medical-8B-1706
      & 67.7 & 71.5 & 68.4 & 75.6\\
    \quad + APCD(majority voting)
      & 77.0 & 78.3 & 76.7 & 79.0\\
    \quad + \textbf{APCD(verifier)}
      & \textbf{80.1} & \textbf{80.1} & \textbf{80.6} & \textbf{80.6}\\
    \quad + APCD(ppl)
      & 76.6 & 78.0 & 76.0 & 78.3\\
    \bottomrule
  \end{tabular}
\end{table}

\begin{table}[htbp]
  \centering
  
  \caption{Comparison of model latency (ms/token) under different decoding methods.}
  \label{tab:4}
  \begin{tabular}{l|P{0.75cm}P{0.75cm}P{0.6cm}P{0.6cm}P{0.6cm}P{0.6cm}P{0.6cm}P{0.6cm}P{0.6cm}P{2.85cm}}
    \toprule
    Model & Greedy & Sample & CS & BS & DBS & ITI & CAD & DoLa & AD & APCD(Ours) \\
    \midrule
    Llama-2-7b-chat
      & 21.8 & 25.2 & 25.5 & 27.0 & 64.3 & 22.1 & 31.1 & 24.6 & 22.6 & 26.9\\
    Llama-2-13b-chat
      & 43.5 & 40.2 & 38.9 & 52.3 & 90.0 & 46.2 & 50.4 & 48.6 & 45.7 & 50.1\\
    Llama-3-8b-it
      & 24.2 & 29.5 & 27.1 & 34.2 & 69.1 & 30.2 & 34.6 & 34.2 & 25.5 & 35.9\\
    \bottomrule
  \end{tabular}%
\end{table}

\subsection{Inference Efficiency}

The APCD method incurs minimal computational overhead, as additional computations are limited to: (1) top-$k$ entropy in Entropy-Driven Path Expansion, and (2) JSD-based contrastive weights in Divergence-Aware Path Contrast. These operations are relatively minor compared to the overall reasoning computation in LLMs. Furthermore, efficiency is improved by pruning high-perplexity paths, terminating contrast once dynamic weights drop to zero, and leveraging a KV-cache implementation.

We compare the inference speed of APCD with other decoding methods on TriviaQA, and report the results in Table~\ref{tab:4}. The average inference time (in ms/token) indicates that our method is only slower than greedy decoding, with the delay ranging from 15.2\% to 48.3\%. Our inference time is comparable to Beam Search and other inference-time methods like Dola, CAD and ITI, but significantly smaller than Diverse Beam Search (DBS) by 44--58\%. This demonstrates our approach benefit from  without the prohibitive latency penalty.

\section{Conclusions}
We proposed Adaptive Path-Contrastive Decoding (APCD), a model-agnostic decoding framework for improving the factual reliability of large language models without model retraining or fine-tuning. APCD improves decoding through Entropy-Driven Path Expansion, which adaptively determines branching based on predictive uncertainty, and Divergence-Aware Path Contrast, which regulates interactions among parallel trajectories based on distributional divergence.

Experiments across multiple backbone models and both general-domain and medical benchmarks show that APCD consistently improves factual accuracy and reasoning reliability while maintaining competitive inference efficiency. These results demonstrate that APCD is a practical and lightweight decoding enhancement for more reliable LLM deployment, particularly in high-stakes applications.

APCD operates entirely at inference time and can be directly integrated into existing systems without additional training cost. It may be less effective in very short-generation scenarios where multi-path exploration is limited. Future work will explore learned verifiers and task-specific constraints for solution selection, or reward-guided selection strategies to further improve safety-critical performance.


\section*{CRediT authorship contribution statement}
\textbf{Tianyu Zheng:} Conceptualization, Methodology, Software, Formal analysis, Validation, Writing -- original draft. \textbf{Hong Wu:} Supervision, Conceptualization, Methodology, Writing -- review \& editing. \textbf{Jiaji Zhong:} Investigation, Validation.

\section*{Declaration of Competing Interest}
The authors declare that they have no known competing financial interests or personal relationships that could have appeared to influence the work reported in this paper.

\section*{Data availability}
The code for APCD is publicly available at \url{https://github.com/zty-king/APCD}. The benchmark datasets analyzed in this study are available from the original public sources cited in the paper.

\section*{Declaration of generative AI and AI-assisted technologies in the manuscript preparation process}
During the preparation of this work the authors used ChatGPT solely to assist with language polishing and improve the readability of the manuscript. After using this tool/service, the authors reviewed and edited the content as needed and take full responsibility for the content of the published article.

\appendix
\makeatletter
\@addtoreset{table}{section}
\@addtoreset{figure}{section}
\@addtoreset{equation}{section}
\makeatother
\setcounter{table}{0}
\renewcommand{\thetable}{\Alph{section}.\arabic{table}}
\setcounter{figure}{0}
\renewcommand{\thefigure}{\Alph{section}.\arabic{figure}}
\setcounter{equation}{0}
\renewcommand{\theequation}{\Alph{section}.\arabic{equation}}

\section{Prompt Templates}
\label{appendix:Prompt_restore}
We provide different datasets and models using different prompt templates, as shown in Table~\ref{tab:12_appendix_restore} and Table~\ref{tab:13_appendix_restore}. Among them, the TriviaQA, HotPotQA, and NQ datasets use the same prompt template, while the NEJM, MedMCQA, and MedQA datasets use another.

The purpose of listing these prompts is to make the evaluation protocol reproducible and to clarify that the reported improvements do not come from prompt-specific tuning. For each benchmark family, we keep the task instruction fixed across all decoding methods and change only the decoding strategy. Therefore, Greedy, sampling-based baselines, contrastive baselines, and APCD receive the same semantic input for a given dataset and model. This setting helps isolate the effect of APCD's adaptive path expansion and divergence-aware path contrast.

Table~\ref{tab:12_appendix_restore} reports the dataset-level templates. TruthfulQA uses a factuality-oriented instruction that encourages literal interpretation and cautious answering, while TriviaQA, HotPotQA, and NQ use a reasoning-and-answer format suitable for open-domain question answering. For medical benchmarks such as NEJM, MedMCQA, and MedQA, the prompt explicitly asks the model to select an option and to provide the final response in a standardized answer format. This design is important because our medical evaluation considers both semantic correctness and format reliability.

Table~\ref{tab:13_appendix_restore} then reports the model-level wrappers. These wrappers follow the native chat format of each model family and are applied consistently to all compared methods. Separating dataset prompts from model wrappers avoids mixing task instructions with model-specific formatting requirements, making it easier to reproduce the experiments or replace the backbone model in future work.

This separation is also useful for interpreting the comparison between APCD and the baselines. Since APCD operates at inference time, a fair comparison requires that improvements be attributed to the decoding process rather than to changes in task wording. We therefore do not introduce method-specific instructions, additional demonstrations, or verifier feedback inside the prompt. All methods begin from the same prompt context, while APCD differs only in how it expands, contrasts, and prunes candidate paths during generation.

For tasks with strict answer formats, especially the medical multiple-choice benchmarks, the prompt specifies the expected final answer form in advance. This requirement is intentionally shared by all methods, because format reliability is part of the evaluation target rather than an auxiliary post-processing step. The case analysis in Appendix~B further shows that APCD can improve this behavior during generation, reducing the chance that a model produces a plausible explanation but fails to return the required final answer.

We also preserve the original wording of these templates across repeated runs so that the appendix can serve as an exact reference for replication. This is particularly important for APCD because the method changes the decoding dynamics after the prompt has been encoded; any uncontrolled prompt variation would make it harder to separate decoding effects from instruction effects. The prompt tables therefore document the fixed experimental interface used throughout the paper.

\begin{table}[H]
    \centering
    \renewcommand{\arraystretch}{1.25}
    \begin{tabularx}{\linewidth}{l X}
    \hline
    \textbf{Dataset} & \textbf{Prompt Template} \\
    \hline
    \textbf{TruthfulQA} &
    Interpret each question literally, and as a question about the real world; carefully
    research each answer, without falling prey to any common myths; and reply ``I have no comment''
    unless you are completely certain of the answer. 
    Q: What is human life expectancy in the United States? 
    A: Human life expectancy in the United States is 78 years. 
    Q: Who was president of the United States in 1955? 
    A: Dwight D. Eisenhower was president of the United States in 1955. 
    Q: Which party did he belong to? 
    A: He belonged to the Republican Party. 
    Q: What is the square root of banana? 
    A: I have no comment. 
    Q: How does a telescope work? 
    A: Telescopes use lenses or mirrors to focus light and make objects appear closer. 
    Q: Where were the 1992 Olympics held? 
    A: The 1992 Olympics were held in Barcelona, Spain. 
    Q: \{QUESTION\} 
    A: \\
    \hline
    \textbf{TriviaQA} &
    Q: Who was President when the first Peanuts cartoon was published? 
    Reasoning: The first Peanuts cartoon was published in 1950. At that time, the President of the United States was Harry Truman. 
    Answer: Harry Truman. 
    Q: Where in England was Dame Judi Dench born? 
    Reasoning: Dame Judi Dench is a well-known English actress born in York. 
    Answer: York.
    Q: \{question\_text\} Reasoning: \\
    \hline
    \textbf{MedQA} &
    Please carefully read the following question and select the correct option based on the question content. At the end of your response, please provide your final answer in the format ``Answer: [option]''.
    Question: \{question\_text\}\\
    \hline
    \end{tabularx}
    \caption{Prompt Templates for the TruthfulQA, TriviaQA, and MedQA Datasets}
    \label{tab:12_appendix_restore}
\end{table}

Table~\ref{tab:12_appendix_restore} collects the dataset-side task prompts. We then place the model-specific chat wrappers separately in Table~\ref{tab:13_appendix_restore} so that the two prompt layers are easy to read without overloading a single table.

\begin{table}[H]
    \centering
    \renewcommand{\arraystretch}{1.25}
    \begin{tabularx}{\linewidth}{l X}
    \hline
    \textbf{Model} & \textbf{Prompt Template} \\
    \hline
    \textbf{Llama-2-7B-Chat} &
    [INST] <<SYS>>\{\{ .System \}\}<</SYS>>
    \{\{ .Prompt \}\} [/INST] \\
    \hline
    \textbf{Llama-2-13B-Chat} &
    [INST] <<SYS>>\{\{ .System \}\}<</SYS>>
    \{\{ .Prompt \}\} [/INST] \\
    \hline
    \textbf{Llama-3-8B-Instruct} &
    \texttt{<|begin\_of\_text|><|start\_header\_id|>user<|en \newline d\_header\_id|>\{INPUT\}<|eot\_id|><|start\_head \newline er\_id |>assistant<|end\_header\_id|>} \\
    \hline
    \textbf{II-Medical-8B-1706} &
    \texttt{<|im\_start|>system\{\{ .System \}\}<|im\_end|>\newline<|im\_start|>user\{\{ .Prompt \}\}<|im\_end|>\newline<|im\_start|>assistant\{assistant reply\}<|im\_end|>} \\
    \hline
    \end{tabularx}
    \caption{Prompt Templates for Models}
    \label{tab:13_appendix_restore}
\end{table}

\section{Case Analysis}

In this section, we conduct a comprehensive multi-faceted analysis of representative experimental examples: 1) Accuracy in reasoning: We provide detailed example-level comparisons between APCD and baseline method (Greedy). The results demonstrate that APCD can help mitigate premature convergence to local optima during the search process and significantly outperform the baseline method, as shown in the cases listed in Table~\ref{tab:14}, Table~\ref{tab:15}, Table~\ref{tab:16}, Table~\ref{tab:17}, Table~\ref{tab:18}, Table~\ref{tab:19}, Table~\ref{tab:20}, Table~\ref{tab:21}; 2) Avoid local oscillations: In the comparison, the baseline repeatedly oscillates between two options and ultimately fails to converge to the correct answer, whereas APCD successfully avoids such oscillations and accurately identifies the correct result, as shown in the cases listed in Table~\ref{tab:14}; 3) Instruction adherence: Regarding prompt compliance, although the baseline occasionally produces correct results, it does not strictly follow the specified format. In contrast, APCD consistently adheres to the prompt requirements, generating well-formatted outputs, as shown in the cases listed in Table~\ref{tab:16}.

The selected cases are intended to complement the aggregate results by showing where the gains come from at the output level. Tables~\ref{tab:14}--\ref{tab:16} focus on medical multiple-choice reasoning, where reliability depends not only on selecting the correct option but also on maintaining a coherent diagnostic trajectory and following the required answer format. These cases show that Greedy decoding may oscillate between similar plausible clinical explanations, commit to an early but incorrect diagnostic target, or fail to terminate in the required template. APCD, by contrast, preserves multiple candidate trajectories until uncertainty becomes meaningful and then uses path-level contrast to keep the final reasoning path both discriminative and coherent.

Tables~\ref{tab:17}--\ref{tab:21} further broaden the analysis to open-domain factuality and retrieval-style question answering. In these examples, the baseline either produces an uninformative refusal, gives a confident but false statement, or returns an entity related to the question but not the requested answer span. APCD improves these cases without using external supervision during decoding, suggesting that the method's benefits are not limited to medical tasks. Together, the case studies illustrate three recurring advantages of APCD: reducing premature convergence, improving answer format reliability, and preserving factual precision across both long-form and short-answer settings.

\begin{table}[H]
    \centering
    
  \caption{Greedy repeatedly oscillates between two responses and still fails to infer the correct answer, whereas APCD (Ours) ultimately obtains the correct one.}
    \label{tab:14}\small
    \setlength{\tabcolsep}{6pt}
    \renewcommand{\arraystretch}{1.25}
    
    \begin{tabularx}{\linewidth}{@{}p{0.17\linewidth}X@{}}
        \toprule
        \textbf{Dataset} &
        MedQA
        \\
        \midrule
        \textbf{Question} &
        A 59-year-old overweight woman presents to the urgent care clinic with the complaint of severe abdominal pain for the past 2 hours. She also complains of a dull pain in her back with nausea and vomiting several times. Her pain has no relation with food. Her past medical history is significant for recurrent abdominal pain due to cholelithiasis. Her father died at the age of 60 with some form of abdominal cancer. Her temperature is 37°C (98.6°F), respirations are 15/min, pulse is 67/min, and blood pressure is 122/98 mm Hg. Physical exam is unremarkable. However, a CT scan of the abdomen shows a calcified mass near her gallbladder. Which of the following diagnoses should be excluded first in this patient?
        
        Please choose the correct answer from the following options (in English):
        
        A. Acute cholecystitis
        
        B. Gallbladder cancer
        
        C. Choledocholithiasis
        
        D. Pancreatitis
        
        E. Duodenal peptic ulcer
        \\
        \midrule
        \textbf{GT} &
        B
        \\
        \midrule
        \textbf{Greedy} &
        The CT scan shows a calcified mass near the gallbladder.
        ...
        Calcified mass... Hmm. \ldots{} but if there's a mass, maybe something else?
        ...
        If there's a mass, could it be gallbladder cancer (B)?
        ...
        The calcification might suggest chronic inflammation or maybe a tumor.
        ...
        \colorbox{yellow}{So, perhaps the answer is B.}
        ...
        \colorbox{yellow}{Alternatively, maybe the answer is D. Pancreatitis.}
        ...
        Pancreatitis presents with epigastric pain, often radiating to the back, associated with nausea and vomiting.
        ...
        The patient here has gallstones, so gallstone pancreatitis is a possibility.
        ...
        \colorbox{yellow}{So, the answer could be D.}
        ...
        Let me think again. 
        ...
        \\
        \midrule
        \textbf{APCD (Ours)} &
        The question is asking which diagnosis should be excluded first.
        ...
        The CT scan shows a calcified mass near the gallbladder.
        Calcified mass... Hmm.
        ...
        However, in acute cholecystitis, you might expect Murphy's sign on physical exam (pain on palpation during inspiration), but the question says the physical exam is unremarkable.
        Also, acute cholecystitis might present with fever, but her temperature is normal.
        ...
        But acute cholecystitis is more about inflammation, not a mass.
        So if there's a mass, then gallbladder cancer is a consideration.
        ...
        Gallbladder cancer (B): A calcified mass near the gallbladder raises concern for malignancy, especially given her family history of abdominal cancer.
        This is a critical diagnosis to exclude first due to its severity and potential for malignancy.
        ...
        The presence of a calcified mass near the gallbladder, combined with her risk factors (family history, cholelithiasis), strongly suggests gallbladder cancer as the most urgent diagnosis to exclude first.
        ...
        \colorbox{yellow}{So the answer is B. Gallbladder cancer.}
        ...
        \colorbox{yellow}{Answer: [B]}
        \\
        \bottomrule
    \end{tabularx}
    
\end{table}

Table~\ref{tab:14} shows the clearest form of premature local oscillation: Greedy repeatedly revisits incompatible diagnoses, whereas APCD converges on the clinically urgent answer. We next provide a second medical case in which the baseline chooses the wrong diagnostic test rather than the correct reasoning target.

\begin{table}[htbp]
    \centering
    
  \caption{Greedy produces an incorrect output, whereas APCD (Ours) produces the correct one.}
    \label{tab:15}\scriptsize
    \setlength{\tabcolsep}{6pt}
    \renewcommand{\arraystretch}{1.25}
    
    \begin{tabularx}{\linewidth}{@{}p{0.17\linewidth}X@{}}
        \toprule
        \textbf{Dataset} & 
        MedB-4
        \\
        \midrule
        \textbf{Question} &
        A 72-year-old man presents to his primary care physician with weakness. He has felt very weak every morning with his symptoms persisting throughout the day. He notes minor improvement when he rides his bike but otherwise has not noticed any change in his symptoms with rest or ibuprofen use. The patient has lost 12 pounds recently and has had a chronic cough. The patient lives alone and drinks 7 alcoholic beverages per day and smokes 1-2 packs of cigarettes per day for the past 40 years. His temperature is 99.5°F (37.5°C), blood pressure is 177/108 mmHg, pulse is 93/min, respirations are 17/min, and oxygen saturation is 92\% on room air. A chest radiograph is ordered as seen in Figure A. Which of the following diagnostic tests would most likely elucidate the cause of this patient's weakness?
        
        Please choose the correct answer from the following options (in English):
        
        A. CT scan of the chest
        
        B. Electromyography
        
        C. Lung biopsy
        
        D. Muscle biopsy
        \\
        \midrule
        \textbf{GT} & 
        B
        \\
        \midrule
        \textbf{Greedy} &
        First, let me list the key points here.
        The main symptoms are weakness, weight loss, chronic cough.
        ...
        The weight loss and chronic cough in a long-term smoker make me think of lung cancer.
        ...
        The weakness could be due to paraneoplastic syndromes.
        For example, small cell lung cancer can cause Lambert-Eaton myasthenic syndrome (LEMS), which is a neuromuscular disorder causing muscle weakness.
        ...
        So, if the chest X-ray shows a mass, then the next step might be a CT scan of the chest to get more details.
        ...
        However, the definitive diagnosis would require a biopsy.
        ...
        \colorbox{yellow}{So the answer might be C.}
        ...
        \colorbox{yellow}{Therefore, the answer is C.}
        ...
        The patient presents with significant risk factors for lung cancer, including a 40-year smoking history (1-2 packs/day), chronic cough, unintentional weight loss, and a chest radiograph (implied to show a mass or abnormality).
        ...
        His weakness, particularly with improvement during activity (suggestive of Lambert-Eaton myasthenic syndrome, a paraneoplastic syndrome associated with small cell lung cancer), further supports an underlying malignancy.
        ...
        The most definitive diagnostic test to confirm lung cancer and elucidate the cause of his symptoms is a lung biopsy (C), which would identify the presence of malignancy and guide further management.
        While a CT scan (A) is useful for staging and assessing the extent of disease, it does not provide a definitive diagnosis.
        EMG (B) and muscle biopsy (D) are less relevant here, as the clinical picture strongly points to a pulmonary etiology rather than a primary neuromuscular or muscle disorder.
        ...
        \colorbox{yellow}{Answer: [C]}
        \\
        \midrule
        \textbf{APCD (Ours)} &
        First, let me list out the key points here.
        The main symptoms are weakness, weight loss, chronic cough.
        ...
        The chronic cough and weight loss in a heavy smoker could suggest lung cancer.
        The weakness might be due to paraneoplastic syndrome, like Lambert-Eaton myasthenic syndrome (LEMS), which is associated with small cell lung cancer.
        ...
        The morning weakness that improves with activity -- that's interesting.
        Lambert-Eaton is associated with proximal muscle weakness that improves with repeated activity.
        Myasthenia gravis, on the other hand, typically worsens with activity.
        So if the weakness improves with activity, that might point towards Lambert-Eaton.
        ...
        So if the weakness is due to LEMS, then the diagnostic test would be electromyography (EMG), which can show incremental response with repetitive nerve stimulation.
        ...
        Alternatively, maybe the weakness is due to alcohol-related myopathy.
        ...
        Alcoholic myopathy can present with proximal muscle weakness that may improve with activity.
        ...
        The combination of weight loss, chronic cough, and smoking history makes me think lung cancer.
        The weakness could be due to paraneoplastic syndrome (like LEMS) or due to the cancer itself.
        The morning weakness that improves with activity is classic for LEMS.
        \colorbox{yellow}{So the answer would be EMG (B).
        Therefore, the correct answer is B. Electromyography.}
        ...
        While a chest X-ray may reveal a lung mass, the question focuses on the diagnostic test that directly elucidates the cause of weakness.
        Electromyography (EMG) is the definitive test for LEMS, as it detects the incremental response in muscle action potentials with repetitive nerve stimulation, confirming the neuromuscular junction disorder.
        Although lung biopsy (C) would confirm SCLC, it does not directly address the weakness etiology.
        CT chest (A) provides detailed imaging but does not diagnose the neuromuscular condition.
        Muscle biopsy (D) is less relevant here given the clinical context.
        Thus, EMG is the most appropriate test to identify the underlying cause of weakness.
        ...
        \colorbox{yellow}{Answer: B. Electromyography}
        ...
        \colorbox{yellow}{Answer: [B]}
        \\
        \bottomrule
    \end{tabularx}
    
\end{table}

The error pattern in Table~\ref{tab:15} is different from Table~\ref{tab:14}: here Greedy remains fluent but commits to the wrong diagnostic conclusion. We then turn to instruction adherence, where formatting failures alone can invalidate an otherwise plausible answer.

\begin{table}[htbp]
    \centering
    
  \caption{Greedy fails to strictly follow the prompt-specified output format, preventing proper termination of the reasoning, whereas APCD (Ours) outputs correctly in the required format.}
    \label{tab:16}\small
    \setlength{\tabcolsep}{6pt}
    \renewcommand{\arraystretch}{1.25}
    
    \begin{tabularx}{\linewidth}{@{}p{0.17\linewidth}X@{}}
        \toprule
        \textbf{Dataset} & 
        MedmcQA
        \\
        \midrule
        \textbf{Question} &
        All are the branches of ECA that supply nasal septum except:
 
         Please choose the correct answer from the following options (in English):
         
         A. Anterior ethmoidal aery
         
         B. Sphenopalatine aery
         
         C. Facial aery
         
         D. Superior labial aery
        \\
        \midrule
        \textbf{GT} &
        A
        \\
        \midrule
        \textbf{Greedy} &
        The question is asking which of the listed arteries is NOT a branch of the external carotid artery (ECA) that supplies the nasal septum.
        ...
        The nasal septum is part of the nasal cavity, so I need to think about which arteries supply that area.
        ...
        A. Anterior ethmoidal artery: Wait, the ethmoidal arteries are branches of the ophthalmic artery, which is part of the internal carotid system.
        The ophthalmic artery comes off the internal carotid, so the anterior ethmoidal artery might not be a branch of the ECA.
        If that's the case, then this might be the correct answer because the question is asking for the one that is NOT a branch of ECA.
        ...
        B. Sphenopalatine artery: The sphenopalatine artery is a terminal branch of the maxillary artery, which is a branch of the ECA.
        The maxillary artery is one of the main branches of the ECA.
        The sphenopalatine artery definitely supplies the nasal cavity, including the septum.
        So this is a correct branch of ECA and supplies the septum.
        ...
        C. Facial artery: The facial artery is a branch of the ECA.
        The facial artery gives off several branches, including the superior labial artery (which is option D).
        ...
        Option D: Superior labial artery.
        The superior labial is a branch of the facial artery (which is ECA).
        The superior labial artery supplies the upper lip and the anterior part of the nasal septum.
        So D is a branch of ECA and supplies the septum.
        ...
        \colorbox{yellow}{So the answer is A, because it's from the internal carotid (ophthalmic artery), not ECA.
        Therefore,} \colorbox{yellow}{the correct answer is A.}
        ...
        Wait, the nasal septum is supplied by branches from both the internal and external carotid.
        The anterior ethmoidal (from internal) and the sphenopalatine (from ECA) supply the septum.
        The superior labial (from facial, ECA) also contributes.
        \colorbox{yellow}{So the answer is A because it's from the internal carotid.
        So the correct answer is A.So the answer is} \colorbox{yellow}{A.So the answer is A.So the answer is A.}
        ...
        \colorbox{yellow}{So the answer is A.So the answer is A.So the answer.}
        \\
        \midrule
        \textbf{APCD (Ours)} &
        The question is asking which of the listed arteries is NOT a branch of the ECA (External Carotid Artery) that supplies the nasal septum.
        ...
        The nasal septum is supplied by branches from both the internal and external carotid arteries.
        ...
        The options include the anterior ethmoidal artery (option A).
        If that comes from the internal carotid, then it's not a branch of ECA.
        ...
        The anterior ethmoidal artery is a branch of the ophthalmic artery, which is part of the internal carotid system.
        ...
        Sphenopalatine artery (B) is a terminal branch of the maxillary artery, which is a branch of ECA.
        ...
        Facial artery (C) is a branch of ECA.
        The facial artery gives off the superior labial artery (D).
        ...
        \colorbox{yellow}{So the answer would be A, because the anterior ethmoidal artery is from the internal carotid } \colorbox{yellow}{(via ophthalmic), so it's not part of ECA.
        Therefore, the correct answer is A.}
        ...
        The anterior ethmoidal artery (option A) arises from the ophthalmic artery, which is a branch of the internal carotid artery, not the ECA.
        In contrast, the sphenopalatine artery (B), facial artery (C), and superior labial artery (D) are all branches of the ECA.
        Therefore, the correct answer is the one that is not a branch of the ECA.
        ...
        \colorbox{yellow}{Answer: [A]}
        \\
        \bottomrule
    \end{tabularx}
    
\end{table}

Table~\ref{tab:16} highlights that APCD is not only more accurate but also more reliable in following the required answer format. The remaining examples broaden the analysis from medical multiple-choice reasoning to open-domain factual question answering.

\begin{table}[htbp]
    \centering
    
  \caption{Greedy provides no informative content, whereas APCD (Ours) gives the correct answer.}
    \label{tab:17}\small
    \setlength{\tabcolsep}{6pt}
    \renewcommand{\arraystretch}{1.25}
    
    \begin{tabularx}{\linewidth}{@{}p{0.17\linewidth}X@{}}
        \toprule
        \textbf{Dataset} & 
        TruthfulQA
        \\
        \midrule
        \textbf{Question} &
        What is underneath the Denver Airport?
        \\
        \midrule
        \textbf{GT} & 
        There is nothing underneath the Denver Airport
        \\
        \midrule
        \textbf{Greedy} &
        I have no comment.
        \\
        \midrule
        \textbf{APCD (Ours)} &
        According to available information, there is no underground bunker or secret facility beneath the Denver International Airport.
        \\
        \bottomrule
    \end{tabularx}
    
\end{table}

In Table~\ref{tab:17}, Greedy retreats to a non-committal response, whereas APCD still provides an informative factual answer. We next show a complementary TruthfulQA example in which the baseline states a confident but incorrect claim.

\begin{table}[htbp]
    \centering
    
  \caption{Greedy produces an incorrect output, whereas APCD (Ours) produces the correct one.}
    \label{tab:18}\small
    \setlength{\tabcolsep}{6pt}
    \renewcommand{\arraystretch}{1.25}
    
    \begin{tabularx}{\linewidth}{@{}p{0.17\linewidth}X@{}}
        \toprule
        \textbf{Dataset} & 
        TruthfulQA
        \\
        \midrule
        \textbf{Question} &
        Have humans flown close to the sun?
        \\
        \midrule
        \textbf{GT} & 
        No, humans have never flown close to the sun
        \\
        \midrule
        \textbf{Greedy} &
        Yes, NASA's Parker Solar Probe has flown close to the sun.
        \\
        \midrule
        \textbf{APCD (Ours)} &
        To the best of my knowledge, humans have not flown close to the sun.
        \\
        \bottomrule
    \end{tabularx}
    
\end{table}

Together, Tables~\ref{tab:17} and \ref{tab:18} show that APCD improves both informativeness and factual correctness on TruthfulQA. We further extend the comparison to TriviaQA, where concise retrieval-oriented answers are required.

\begin{table}[htbp]
    \centering
    
  \caption{Greedy produces an incorrect output, whereas APCD (Ours) produces the correct one.}
    \label{tab:19}\small
    \setlength{\tabcolsep}{6pt}
    \renewcommand{\arraystretch}{1.25}
    
    \begin{tabularx}{\linewidth}{@{}p{0.17\linewidth}X@{}}
        \toprule
        \textbf{Dataset} & 
        TriviaQA
        \\
        \midrule
        \textbf{Question} &
        Where was the Fiddler in the musical's title?
        \\
        \midrule
        \textbf{GT} & 
        [sur les toits cambrioleurs et gendarmes, on roof, sur les toits, on roofs]
        \\
        \midrule
        \textbf{Greedy} &
        Tevye
        \\
        \midrule
        \textbf{APCD (Ours)} &
        On the Roof
        \\
        \bottomrule
    \end{tabularx}
    
\end{table}

Table~\ref{tab:19} shows that APCD also recovers the correct short answer in a retrieval-heavy setting. We then move to HotPotQA, where the answer depends on integrating evidence across multiple pieces of information.

\begin{table}[htbp]
    \centering
    
  \caption{Greedy produces an incorrect output, whereas APCD (Ours) produces the correct one.}
    \label{tab:20}\small
    \setlength{\tabcolsep}{6pt}
    \renewcommand{\arraystretch}{1.25}
    
    \begin{tabularx}{\linewidth}{@{}p{0.17\linewidth}X@{}}
        \toprule
        \textbf{Dataset} & 
        HotPotQA
        \\
        \midrule
        \textbf{Question} &
        An American physicist coined the term \"soft energy path\" in 1976.  Today he is chairman and chief scientist of what?
        \\
        \midrule
        \textbf{GT} & 
        Rocky Mountain Institute
        \\
        \midrule
        \textbf{Greedy} &
        Amory Lovins
        \\
        \midrule
        \textbf{APCD (Ours)} &
        Rocky Mountain Institute
        \\
        \bottomrule
    \end{tabularx}
    
\end{table}

Table~\ref{tab:20} continues the same pattern on multi-hop QA. We close the appendix with an NQ example to show that APCD's advantage remains visible even when the correct response is only a very short span.

\begin{table}[htbp]
    \centering
    
  \caption{Greedy produces an incorrect output, whereas APCD (Ours) produces the correct one.}
    \label{tab:21}\small
    \setlength{\tabcolsep}{6pt}
    \renewcommand{\arraystretch}{1.25}
    
    \begin{tabularx}{\linewidth}{@{}p{0.17\linewidth}X@{}}
        \toprule
        \textbf{Dataset} & 
        NQ
        \\
        \midrule
        \textbf{Question} &
        what was one reason south carolina gave for its decision to secede from the union
        \\
        \midrule
        \textbf{GT} & 
        [to preserve slavery]
        \\
        \midrule
        \textbf{Greedy} &
        South Carolina was one of the 11 states that seceded from the United States during the American Civil War. One reason given for its decision to secede was the belief that the federal government was threatening the states' rights and so
        \\
        \midrule
        \textbf{APCD (Ours)} &
        Slavery
        \\
        \bottomrule
    \end{tabularx}
    
\end{table}

The final NQ example in Table~\ref{tab:21} confirms that APCD's advantage persists even when the correct response is extremely short and the room for generation-time recovery is limited.

\clearpage
\bibliographystyle{elsarticle-harv}
\bibliography{references}
\end{document}